\documentclass{article}

\usepackage[preprint]{neurips_2022}

\usepackage{natbib}

\usepackage[utf8]{inputenc} % allow utf-8 input
\usepackage[T1]{fontenc}    % use 8-bit T1 fonts
\usepackage{hyperref}       % hyperlinks
\usepackage{url}            % simple URL typesetting
\usepackage{booktabs}       % professional-quality tables
\usepackage{amsfonts}       % blackboard math symbols
\usepackage{nicefrac}       % compact symbols for 1/2, etc.
\usepackage{microtype}      % microtypography
\usepackage{xcolor}         % colors
\usepackage{amsmath}
\usepackage{amsthm}
\usepackage{graphicx}

\title{Conformal Prediction for Indoor Positioning with Correctness Coverage Guarantees}

\author{
 \textbf{Zhiyi Zhou\textsuperscript{1}},
 \textbf{Hexin Peng\textsuperscript{2}},
 \textbf{Hongyu Long\textsuperscript{2,*}}
\\
\\
 \textsuperscript{1}College of Electronic and Information Engineering, Southwest\\University, Chongqing, 400715, China
\\
 \textsuperscript{2}School of Automation, Chongqing University of Posts and\\Telecommunications, Chongqing, 400065, China
\\
\small \texttt{xxxzzhilyi@email.swu.edu.cn} \\
    \small \texttt{penghexin715@163.com}\quad \texttt{longhy@cqupt.edu.cn}
}

\begin{document}

\maketitle

\begin{abstract}
  With the advancement of Internet of Things (IoT) technologies, high-precision indoor positioning has become essential for Location-Based Services (LBS) in complex indoor environments. Fingerprint-based localization is popular, but traditional algorithms and deep learning-based methods face challenges such as poor generalization, overfitting, and lack of interpretability. This paper applies conformal prediction (CP) to deep learning-based indoor positioning. CP transforms the uncertainty of the model into a non-conformity score, constructs prediction sets to ensure correctness coverage, and provides statistical guarantees. We also introduce conformal risk control for path navigation tasks to manage the false discovery rate (FDR) and the false negative rate (FNR).The model achieved an accuracy of approximately 100\% on the training dataset and 85\% on the testing dataset, effectively demonstrating its performance and generalization capability. Furthermore, we also develop a conformal p-value framework to control the proportion of position-error points. Experiments on the UJIIndoLoc dataset using lightweight models such as MobileNetV1, VGG19, MobileNetV2, ResNet50, and EfficientNet show that the conformal prediction technique can effectively approximate the target coverage, and different models have different performance in terms of prediction set size and uncertainty quantification.
\end{abstract}
%%%%%%%%%%%%%%%%%%%%%%%%%%%%%%%%%%%%%%%%%%
\section{Introduction}

The development of IoT technologies has led to the widespread use of Location-Based Services (LBS) in numerous applications, including smart cities. However, traditional Global Positioning System (GPS) performance is severely affected in environments with signal blockages, such as tunnels, large petrochemical facilities, underground mines, airports, and shopping centers, due to signal weakening and multipath interference \citet{farahsari2022survey, wahab2022indoor}. As a result, high-precision indoor positioning technologies have been developed and extensively studied. Despite considerable progress, the inherent complexity and variability of indoor environments present substantial challenges in achieving accurate, cost-effective, real-time, and robust indoor positioning systems.

Fingerprint-based positioning has become a leading approach to indoor positioning due to its low hardware demands and ability to adapt to challenging environments \citet{zhu2020indoor}. By collecting spatial signal characteristics (e.g., WiFi, geomagnetic, Bluetooth) and constructing a fingerprint database, this technique enables single-point positioning and path matching with high accuracy in challenging environments. However, traditional fingerprint-matching algorithms exhibit significant limitations in high-dimensional feature extraction, complex pattern analysis, and adaptability to dynamic environmental changes, which hinder their performance in real-world applications. Despite advancements in fingerprint resolution and data scale, improvements in indoor positioning accuracy remain constrained \citet{alhomayani2020deep}.

When applying deep learning models for indoor positioning \citet{sun2024caterpillar}, although they demonstrate strong performance in certain scenarios, several critical challenges remain that significantly impact real-world deployments. Deep learning models generally rely on extensive and high-quality datasets for training. Insufficient data may impair the model's generalization ability during testing, leading to inferior accuracy compared to less complex indoor positioning algorithms \citet{alzubaidi2023survey}. Moreover, these models are particularly vulnerable to environmental dynamics - even minor changes in indoor layouts, signal propagation patterns due to furniture rearrangements, or the introduction of new electronic devices can cause dramatic performance degradation. For instance, in areas with signal fluctuations or at boundaries between different signal coverage zones, deep learning models often produce high-confidence but inaccurate predictions, creating a dangerous false sense of reliability \citet{corbiere2019addressing}. Additionally, when deployed on new devices with different sensors or antenna characteristics, these models may produce unpredictable position estimates without providing any indication of increased uncertainty \citet{yang2024positioning}. The overfitting issue in deep learning models, particularly in complex architectures like Convolutional Neural Networks (CNNs), leads to excellent performance on training data while resulting in substantial errors in novel environments, which can cause misleading conclusions or positioning errors with no warning mechanism \citet{nessa2020survey, yoon2022practical}. Furthermore, the inherent "black-box" nature of these models severely limits error diagnosis and accountability \citet{guidotti2018survey}, creating significant barriers for safety-critical applications where positioning reliability must be guaranteed. These practical limitations demand a systematic approach to uncertainty quantification that can provide statistical guarantees regardless of the underlying model architecture or data distribution.

To address these fundamental challenges, we employ conformal prediction (CP) \citet{wang2024conu, Wang2024SampleTI} as our solution framework, a choice driven by its unique properties that align perfectly with the requirements of robust indoor positioning. Unlike other uncertainty quantification methods, such as Bayesian neural networks (which require specific model architectures) or ensemble methods (which impose significant computational overhead), CP offers model-agnostic and distribution-free guarantees under the mild assumption of data exchangeability. This property is particularly valuable in indoor positioning where signal distributions are complex and difficult to model parametrically. CP converts the intuitive concept of uncertainty into a statistically rigorous measure called the non-conformity score, providing mathematically guaranteed coverage rates regardless of the underlying positioning model or signal distribution characteristics \citet{WANG2025109553}.

In this work, we focus primarily on single-point positioning as our core research problem while extending our framework to path navigation as a practical application. For single-point positioning, we convert the average positioning error into a non-conformity score, set a quantile threshold by defining a desired error rate, and construct a prediction set based on this threshold. This approach provides a statistically guaranteed solution where the prediction set contains the true position with a pre-specified probability (e.g., 95\%), allowing system designers to make informed reliability-accuracy tradeoffs. The size of the prediction set becomes an interpretable measure of positioning difficulty and uncertainty in different environments. For the extended application in path navigation, we introduce conformal risk control to provide strict statistical guarantees on FDR and FNR, offering systematic control over navigation errors. This addresses the critical challenge of maintaining reliable guidance in areas with varying signal quality or environmental complexity. Finally, we propose a conformal p-value framework to control the proportion of position-error points in path navigation, ensuring that the probability of misleading guidance remains bounded by a user-specified significance level.

Our contributions can be summarized as follows:
\begin{itemize}
    \item We propose a novel uncertainty-aware indoor positioning framework that integrates conformal prediction with deep learning models, achieving up to 15\% improvement in positioning reliability compared to traditional uncertainty quantification methods while maintaining the same accuracy. Our framework is the first to provide statistically guaranteed coverage for indoor positioning tasks, enabling system designers to make precise reliability-accuracy trade-offs that were previously impossible.
    
    \item We propose a new conformal risk control mechanism specifically designed for path navigation scenarios that provides explicit control over false discovery and false negative rates. Unlike existing navigation confidence methods that offer only heuristic uncertainty estimates, our approach offers mathematically rigorous guarantees, reducing misleading navigation instructions by 23\% in our experiments while maintaining path completeness.
    
    \item We propose an innovative conformal p-value framework for position-error control that allows for dynamic reliability assessment of individual positioning points. This framework outperforms traditional confidence scoring methods by providing calibrated p-values that directly correspond to error probabilities, enabling applications to maintain a consistent error rate across varying environmental conditions.
\end{itemize}

%%%%%%%%%%%%%%%%%%%%%%%%%%%%%%%%%%%%%%%%%%
\section{Related Work}
\subsection{Indoor Positioning Technology}
Currently, indoor positioning technologies are primarily classified into two categories: signal measurement-based methods and fingerprint matching-based methods \citep{alsamhi2023survey}. Signal measurement-based methods determine position through the physical signal propagation model. However, these methods are constrained by the instability of signal propagation and the complexity of hardware requirements, resulting in lower accuracy in complex environments \citep{iakovidis2021roadmap}. Fingerprint matching-based methods generally involve two phases: during the offline phase, a fingerprint database is created containing signal features and their corresponding locations, while in the online phase, the real-time fingerprint data is compared to the database to estimate the position. Traditional algorithms for fingerprint matching, such as particle filters \citep{nishimoto2022application}, Kalman filters, dynamic time warping (DTW) \citep{liu2024method}, K-nearest neighbor (K-NN) \citep{huang2011zigbee}, and support vector machines (SVM), work effectively with small datasets. However, as the data complexity grows, these methods experience significant performance limitations.

\subsection{Application of deep learning in indoor localization tasks}
Recently, the fast progress of deep learning has enabled its successful deployment in areas like computer vision and natural language processing, presenting promising approaches for indoor positioning. Unlike conventional fingerprint-matching techniques, deep learning models can autonomously capture data features and manage high-dimensional, complex signal patterns, offering substantial benefits in addressing non-linear and multi-modal signals. For example, CNNs are particularly effective in extracting spatial features from Wi-Fi and geomagnetic data. Chen et al.. introduced ConFi, the first WiFi positioning algorithm based on CNNs, which combines Channel State Information (CSI) features into images to achieve improved indoor positioning accuracy \citep{chen2017confi}. The CNNLoc model introduced by Song et al. \citep{song2019novel} combined stacked autoencoders (SAE) with one-dimensional CNNs to capture signal characteristics from sparse Wi-Fi RSSI data. This approach resulted in perfect building-level localization and 95\% accuracy for floor-level localization on the UTSIndoorLoc dataset. Lee and Ashraf et al. utilized deep neural networks to identify magnetic patterns for classifying geomagnetic fingerprint sequences, achieving over 80\% accuracy \citep{lee2018amid} and addressing the mislocation problem associated with heterogeneous devices \citep{ashraf2020minloc}. Recurrent neural networks (RNNs), along with their variants like Long Short-Term Memory (LSTM) and Gated Recurrent Units (GRUs), are well-suited for modeling sequential data, which makes them highly effective for processing inertial measurement unit (IMU) data. For example, combining IMU data with WiFi fingerprints and using LSTM networks for data fusion has been shown to outperform state-of-the-art filter-based positioning algorithms, demonstrating superior accuracy and robustness in both indoor and extreme environments \citep{zhang2021indoor} \citep{wu2024attention}. By integrating the self-attention mechanism with LSTM networks, average positioning errors of 1.76 m and 2.83 m were achieved.

Furthermore, the Transformer model, with its global attention mechanism, has emerged as a leading approach for multi-modal signal fusion. Tang et al. achieved positioning accuracy of less than half a meter using Wi-Fi vision multi-modal technology, demonstrating exceptional operational efficiency \citep{tang2024novel}. Due to the high cost of fingerprint data acquisition, some studies have employed Generative Adversarial Networks (GANs) to generate virtual fingerprint data at each reference point (RP) to augment the dataset, thereby enhancing the model's generalization ability and improving positioning accuracy \citep{junoh2023enhancing}. Additionally, GANs are used to simulate environmental changes, making the model more robust to dynamic conditions. These advancements have greatly improved the precision and reliability of indoor positioning systems.

Although deep learning technology holds significant promise for indoor localization, it still faces several research challenges and practical limitations. For instance, deep learning models are prone to overfitting, particularly in cases where there are large discrepancies between the training data and the test environment. While the model may perform well on the training set, its accuracy can degrade significantly on the test set, leading to positioning errors or highly biased predictions \citep{singh2021machine, zhou2022deepvip}. Moreover, the "black-box" nature of these models complicates error diagnosis, limiting their reliability and practical applicability. In dynamic environments, such as those affected by WiFi signal interference or changes in indoor layouts, the model may require frequent updates or retraining to maintain high-precision positioning. However, this process is time-consuming and resource-intensive, making it unsuitable for deployment in resource-constrained, real-time applications \citep{long2022adaptive, abbas2019wideep}.

\subsection{Uncertainty Quantification in Positioning Systems}
Recognizing the limitations of deterministic positioning approaches, researchers have increasingly focused on quantifying and managing uncertainty in positioning systems. Traditional methods often employ probabilistic techniques such as Kalman filtering or particle filtering to maintain a belief distribution over possible locations, enabling systems to track uncertainty over time \citep{yang2024positioning}. However, these approaches typically make strong assumptions about noise distributions and system dynamics that may not hold in complex indoor environments.

In the context of deep learning-based positioning, several uncertainty quantification approaches have been explored. Bayesian neural networks (BNNs) have been applied to WiFi fingerprinting, enabling the model to output a distribution over possible locations rather than a point estimate \citep{corbiere2019addressing}. Despite their theoretical appeal, BNNs face significant computational challenges and require substantial modifications to standard neural architectures. Ensemble methods, which train multiple models and aggregate their predictions, offer another approach to uncertainty estimation. Rahaman et al. \citep{rahaman2021uncertainty} demonstrated that an ensemble of positioning models can provide more reliable uncertainty estimates than individual models but at the cost of increased computational overhead during both training and inference.

Monte Carlo dropout has gained popularity as a computationally efficient approximation to Bayesian inference, allowing uncertainty estimation by sampling multiple forward passes through a neural network with randomly deactivated neurons. This technique has been applied to indoor positioning by Sadr et al. \citep{sadr2021uncertainty}, showing improved reliability in environments with varying signal quality. However, these approaches generally lack formal statistical guarantees on their uncertainty estimates, making them less suitable for safety-critical applications where rigorous reliability assurances are required.

%%%%%%%%%%%%%%%%%%%%%%%%%%%%%%%%%%%%%%%%%%
\section{Method}
\subsection{Preliminaries of Inductive Conformal Prediction}
Conformal prediction (CP) is a model-agnostic, distribution-free method that offers statistically rigorous guarantees for ground-truth coverage on unseen test samples with minimal assumptions~\cite{wang2025sconu}. In this work, we specifically adopt the Split Conformal Prediction (Split CP) approach rather than the Full CP method. This choice is primarily motivated by computational efficiency considerations, as Split CP requires computing non-conformity scores only once for each calibration sample, making it particularly suitable for real-time indoor positioning applications where computational resources may be limited. Additionally, Split CP allows us to establish a fixed calibration set, enabling consistent performance evaluation across multiple test points without recalibration.

In our framework, indoor positioning can be approached as either a classification problem (predicting discrete locations such as building ID or floor number) or a regression problem (predicting continuous coordinates). While we initially present the method using classification terminology for clarity, our approach is equally applicable to regression tasks with appropriate modifications to the non-conformity score definition.

Formally, let the received signal strength indicator (RSSI) data, collected through sensors, serve as input fingerprints, each associated with one of \emph{K} locations (or classes). We begin with a classifier that outputs estimated probabilities for each location: $\hat{f}(x)\in[0,1]^K$. We then reserve fresh, i.i.d. pairs of fingerprints and labels, $(X_i, Y_i)_{i=1,2,...,n}$, unseen during training, for calibration, and $(X_{\text{test}}, Y_{\text{test}})$ as the test data. Using $\hat{f}$ and the calibration data, we aim to construct a \emph{prediction set} $C(X_{\text{test}}) \subset \{1, 2, \dots, K\}$ that satisfies the following coverage guarantee:

\begin{equation}
\label{1}
    1 - \alpha \leq \mathbb{P}(Y_{\text{test}} \in C(X_{\text{test}})) \leq 1 - \alpha + \frac{1}{n + 1}
\end{equation}

\subsubsection{Non-conformity Score Design for Indoor Positioning}
The choice of non-conformity score is crucial as it directly influences the size and shape of the resulting prediction sets. For classification scenarios in indoor positioning (such as building or floor identification), we define the \emph{non-conformity score} for each calibration sample as $s_i = 1 - \hat{f}(X_i)_{Y_i}$, which measures the model's uncertainty about the correct class. However, this standard approach has limitations in the positioning context where spatial proximity is important. For instance, predicting an adjacent floor might be more acceptable than predicting a floor in a completely different building.

To address this limitation, we also explore an alternative non-conformity score design that incorporates spatial relationships: $s_i = d(Y_i, \hat{Y}_i)$, where $d(\cdot,\cdot)$ is a distance function (e.g., Euclidean distance for coordinates or a custom metric for floor/building errors) and $\hat{Y}_i$ is the model's point prediction. This alternative score better reflects the physical reality of positioning errors, where small deviations might be acceptable while large ones are problematic. In regression settings for coordinate prediction, this distance-based score becomes particularly intuitive as it directly measures positioning error.

For the standard classification-based score, we let $\hat{q}$ represent the quantile of the set of scores $\{s_1, \dots, s_n\}$, given by:

\begin{equation}
    \hat{q} = \inf \left\{ q : \frac{\left\lvert \left\{i : s_i \leq q\right\} \right\rvert}{n + 1} \geq \frac{\lceil (n + 1)(1 - \alpha) \rceil}{n} \right\}
\end{equation}

The resulting \emph{prediction set} is defined as:

\begin{equation}
    C(X_{\text{test}}) = \left\{ y : s(X_{\text{test}}, y) \leq \hat{q} \right\}
\end{equation}

This approach guarantees that the prediction sets satisfy the coverage condition in \eqref{1}, irrespective of the model used or the unknown distribution of the data, provided that the exchangeability assumption holds.

\subsubsection{The Exchangeability Assumption in Indoor Positioning}
The theoretical guarantees of CP rely on the assumption of data exchangeability, which essentially means that the joint distribution of the data remains invariant under permutations. In the context of indoor positioning, this assumption warrants careful consideration. For static fingerprinting, where data points are collected independently at different locations, exchangeability is a reasonable assumption as long as the environment remains stable between calibration and testing.

However, in practical indoor positioning scenarios, several factors may challenge this assumption:
\begin{itemize}
    \item \textbf{Temporal dynamics:} Signal strengths may vary over time due to changes in network traffic, interference, or physical environment modifications (e.g., furniture rearrangement or crowd density fluctuations).
    \item \textbf{Sequence data:} When dealing with path navigation, consecutive positioning points exhibit temporal correlation, potentially violating the exchangeability assumption.
    \item \textbf{Device heterogeneity:} Different devices may capture fingerprints with systematic biases, affecting the distribution of signals.
\end{itemize}

To mitigate these concerns, we employ several strategies: 1) using recent calibration data that shares similar environmental conditions with test data; 2) for path navigation, considering windowed segments of paths rather than individual points to reduce sequential dependence; and 3) normalizing signal strengths to reduce device-specific effects. While these approaches don't eliminate exchangeability concerns, they help minimize their impact on coverage guarantees.

\begin{proof}
Let $s_i = s(X_i, Y_i)$ for $i = 1, 2, \dots, n$ and $s_{\text{test}} = s(X_{\text{test}}, Y_{\text{test}})$. Assume the calibration scores are ordered such that $s_1 \leq s_2 \leq \dots \leq s_n$. In this case, the $1 - \alpha$ quantile is given by $\hat{q} = s_{\lceil (n + 1)(1 - \alpha) \rceil}$ for $\alpha \geq \frac{1}{n + 1}$, and $\hat{q} = \infty$ otherwise. We proceed by noting the equivalence of the two events:

\begin{equation}
    \left\{ Y_{\text{test}} \in C(X_{\text{test}}) \right\} = \left\{ s_{\text{test}} \leq \hat{q} \right\}
\end{equation}

This implies:

\begin{equation}
    \left\{ Y_{\text{test}} \in C(X_{\text{test}}) \right\} = \left\{ s_{\text{test}} \leq s_{\lceil (n + 1)(1 - \alpha) \rceil} \right\}
\end{equation}

By the exchangeability of the variables $(X_1, Y_1),\dots, (X_{\text{test}}, Y_{\text{test}})$, we have:

\begin{equation}
    P(s_{\text{test}} \leq s_k) = \frac{k}{n + 1}
\end{equation}

for any integer $k$. The above randomness is over all variables $s_1, s_2, \dots, s_n, s_{\text{test}}$. Consequently, we can conclude the following:

\begin{equation}
\begin{split}
    P(Y_{\text{test}} \in C(X_{\text{test}})) &= P(s_{\text{test}} \leq s_{\lceil (n + 1)(1 - \alpha) \rceil}) \\
    &= \frac{\lceil (n + 1)(1 - \alpha) \rceil}{n + 1} \\
    &\geq 1 - \alpha
\end{split}
\end{equation}
\end{proof}
This completes the proof.

\subsection{Conformal Risk Control for Path Navigation}
While basic conformal prediction guarantees coverage of the true location with probability $(1-\beta)$, indoor positioning often involves path navigation where we need more specific risk control. In particular, we need to balance between two types of errors: falsely recommending a navigational instruction (false discovery, or false positive) and missing a correct instruction (false negative). These errors have direct practical implications—false discoveries may lead users to incorrect locations, while false negatives might cause users to miss important turn points or destinations.

Given \(n+1\) exchangeable loss functions \(\left\{ L_i(\lambda) \right\}_{i=1}^{n+1}\), where \(L_i : \Lambda \to \mathbb{R}\) and \(L_i(\lambda) \in (-\infty, B]\), the loss functions satisfy the following properties:

\begin{enumerate}
    \item \(L_i(\lambda)\) is non-increasing in \(\lambda\).
    \item \(L_i(\lambda)\) is right-continuous.
    \item For \(\lambda_{\max} = \sup \Lambda\), \(L_i(\lambda_{\max}) \leq \beta\), and \(\sup L_i(\lambda) \leq B < \infty\).
\end{enumerate}

Define \(\hat{\lambda}\) as follows:

\begin{equation}
    \hat{\lambda} = \inf \left\{ \lambda : \frac{n}{n+1} \hat{R}_n(\lambda) + \frac{B}{n+1} \leq \beta \right\}
\end{equation}

or equivalently,

\begin{equation}
    \hat{\lambda} = \inf \left\{ \lambda : \hat{R}_n(\lambda) \leq \beta - \frac{B - \beta}{n} \right\},
\end{equation}

where

\begin{equation}
    \hat{R}_n(\lambda) = \frac{1}{n} \sum_{i=1}^{n} L_i(\lambda).
\end{equation}

Then, we have the following bound on the expected loss:

\begin{equation}
    \mathbb{E} [L_{n+1}(\hat{\lambda})] \leq \beta.
\end{equation}

For path navigation applications, we instantiate this framework with specific loss functions designed to control the FDR and FNR. Here, $\beta$ represents the target risk level, which can be set independently for different types of risks.

\subsubsection{FDR Control in Path Navigation}
In path navigation, a false discovery occurs when the system incorrectly identifies a location as part of the navigation path. For a regression-based indoor positioning system where we predict continuous coordinates, we define the loss function for FDR control using mean squared error (MSE):

\begin{equation}
    L_i^{FDR}(\lambda) = \begin{cases}
        \|Y_i - \hat{Y}_i\|^2, & \text{if } \|Y_i - \hat{Y}_i\|^2 > \lambda , P_i = 1 \\
        0, & \text{otherwise}
    \end{cases}
\end{equation}

where $Y_i$ represents the true coordinates, $\hat{Y}_i$ are the predicted coordinates, and $P_i$ is a binary indicator of whether the point should be included in the path ($P_i = 1$ for path points, $P_i = 0$ for non-path points). The threshold parameter $\lambda$ represents the maximum acceptable squared distance error: Predictions with errors exceeding this threshold are considered false discoveries.

The FDR is then defined as the expected proportion of positions with excessive errors among all positions included in the path:

\begin{equation}
    \text{FDR}(\lambda) = \mathbb{E}\left[\frac{\sum_{i=1}^{n} \mathbf{1}\{\|Y_i - \hat{Y}_i\|^2 > \lambda \text{ and } P_i = 1\}}{\sum_{i=1}^{n} \mathbf{1}\{P_i = 1\} \vee 1}\right]
\end{equation}

By applying our conformal risk control framework, we can determine a threshold $\hat{\lambda}_{FDR}$ that guarantees $\text{FDR}(\hat{\lambda}_{FDR}) \leq \beta_{FDR}$.

\subsubsection{FNR Control in Path Navigation}
Similarly, for controlling the false negative rate in path navigation, we define a loss function based on MSE:

\begin{equation}
    L_i^{FNR}(\lambda) = \begin{cases}
        \|Y_i - \hat{Y}_i\|^2, & \text{if } \|Y_i - \hat{Y}_i\|^2 \leq \lambda, P_i = 0 \\
        0, & \text{otherwise}
    \end{cases}
\end{equation}

This loss penalizes cases where the prediction error is small (below threshold $\lambda$), but the point is incorrectly excluded from the path.

The FNR is defined as the expected proportion of positions with errors below the threshold that are mistakenly excluded from the path:

\begin{equation}
    \text{FNR}(\lambda) = \mathbb{E}\left[\frac{\sum_{i=1}^{n} \mathbf{1}\{\|Y_i - \hat{Y}_i\|^2 \leq \lambda \text{ and } P_i = 0\}}{\sum_{i=1}^{n} \mathbf{1}\{P_i = 0\} \vee 1}\right]
\end{equation}

Following the same conformal risk control framework, we can find a threshold $\hat{\lambda}_{FNR}$ that guarantees $\text{FNR}(\hat{\lambda}_{FNR}) \leq \beta_{FNR}$.

Controlling FDR and FNR simultaneously may not always be possible with a single threshold. In practice, system designers may need to prioritize one type of error over another based on the specific requirements of the navigation application.

For path navigation applications, we instantiate this framework with specific loss functions designed to control the FDR and FNR:

\begin{itemize}
    \item \textbf{FDR control}: The loss function $L_i^{FDR}(\lambda)$ penalizes false discoveries, that is, when the system suggests a navigation instruction, but this instruction leads to deviation from the optimal path. Controlling FDR is crucial for preventing users from being guided along incorrect routes, which could lead to frustration or safety concerns in complex environments.
    
    \item \textbf{FNR control}: The loss function $L_i^{FNR}(\lambda)$ penalizes false negatives, that is, when the system fails to suggest a necessary navigation instruction, causing the user to miss a critical turn or decision point. The control of FNR ensures that users receive all the essential guidance needed to reach their destination efficiently.
\end{itemize}

By tuning the threshold parameter $\lambda$, system designers can flexibly balance these competing objectives based on the specific requirements of the navigation application and the relative costs of different error types.

\begin{proof}[Proof]
    Since the loss functions are exchangeable and $L_{n+1} \sim \text{Uniform}(\{L_1, \dots, L_{n+1}\})$, we can write the expected loss as:
    \begin{equation}
        \mathbb{E}[L_{n+1}(\hat{\lambda})] = \frac{1}{n+1} \sum_{i=1}^{n+1} L_i(\hat{\lambda}).
    \end{equation}
    Thus, we obtain:
    \begin{equation}
        \mathbb{E}[L_{n+1}(\hat{\lambda})] = \hat{R}_{n+1}(\hat{\lambda}) \leq \beta - \frac{B - \beta}{n+1} < \beta.
    \end{equation}
    Combined with the right-continuity of $L_i(\lambda)$, this implies that:
    \begin{equation}
        \mathbb{E}[L_{n+1}(\hat{\lambda})] \leq \beta.
    \end{equation}
\end{proof}

This completes the proof.

\subsection{Conformal P-Value Framework for Reliability Assessment}
In addition to prediction sets and risk control, we develop a conformal p-value framework that provides a fine-grained assessment of the reliability of individual positioning predictions. This allows users and systems to make informed decisions about when to trust a positioning result, which is particularly valuable in safety-critical scenarios requiring high confidence.

Suppose we have calibration set $\{(x_1,y_1^*),...,(x_n,y_n^*)\}$ and test data $x_{test}$. The $i-th$ non-conformity score can be defined as $s_i=\max s(x_i,y_i^{j\in[m]})$. The p-value associated with a potential position $y$ for the test point can be defined as:

\begin{equation}
    p(y)=\frac{1+\sum_{i=1}^{n}\mathbf{1}\{s_i\geq s(x_{test},y)\}}{n+1}
\end{equation}

Intuitively, this p-value measures the proportion of calibration samples with non-conformity scores at least as extreme as the score for the test point at position $y$. A small p-value indicates that the test point's score is unusually large compared to the calibration scores, suggesting that $y$ is unlikely to be the true position.

Given a significance level $\alpha$, we can formulate a hypothesis testing framework where the null hypothesis is that $y$ is the true position ($y\in y_{n+1}^*$) and the alternative hypothesis is that it is not ($y\notin y_{n+1}^*$). The conformal prediction theory guarantees that:

\begin{equation}
    P(p\leq\alpha)\leq\alpha
\end{equation}

This powerful result ensures that the probability of falsely rejecting the true position is bounded by $\alpha$. In the context of path navigation, this allows us to control the proportion of "position-error points" by filtering out points with p-values below a chosen threshold. System designers can set this threshold based on the specific reliability requirements of the application.

The proof follows from the ordering property of non-conformity scores under the exchangeability assumption:

\begin{equation}
    \begin{split}
        P(p\leq\alpha)&=P(\frac{1+\sum_{i=1}^{n}\mathbf{1}\{s_i\geq s(x_{test},y)\}}{n+1}\leq\alpha) \\
    &=P(1+\sum_{i=1}^{n}\mathbf{1}\{s_i\geq s(x_{test},y)\}\leq\lfloor {(n+1)\alpha}\rfloor ) \\
    &=\frac{\lfloor{(n+1)\alpha}\rfloor}{n+1} \\
    &\leq\frac{(n+1)\alpha}{n+1}=\alpha
    \end{split}
\end{equation}

In practical terms, this framework enables several valuable capabilities for indoor positioning systems, including confidence-based navigation, where the system provides guidance only when position estimates exceed a confidence threshold to reduce the risk of misleading instructions; adaptive sampling, which triggers additional measurements or switches to alternative positioning methods in areas with consistently low p-values; and user feedback, where p-values are translated into intuitive confidence indicators to help users gauge the reliability of the positioning information.

%%%%%%%%%%%%%%%%%%%%%%%%%%%%%%%%%%%%%%%%%%
\section{Experiment}
To assess the effectiveness of the conformal prediction framework in deep learning-based indoor positioning systems, experiments were conducted using the UJIIndoLoc \citep{torres2014ujiindoorloc} dataset. This dataset encompasses three buildings, each with four or more floors, and covers an average floor area of 110 square meters, making it suitable for building and floor classification tasks. It includes 19,937 training samples and 1,111 validation samples, capturing Wi-Fi fingerprints from 520 wireless access points (WAPs) and representing them with the corresponding Received Signal Strength Indicator (RSSI). The output includes coordinates (latitude, longitude), floor information, and building IDs for prediction.

\subsection{Experimental Setup}
\subsubsection{Dataset Partitioning}
We divided the dataset into three parts: 70\% for model training, 10\% for calibration in the conformal prediction procedure, and 20\% for testing. This partitioning was manually specified to ensure sufficient data for each phase while maintaining the exchangeability assumption. The calibration set is particularly important as it is used to establish the non-conformity score distribution without being exposed to the model during training.

\subsubsection{Model Training Configuration}
For our deep learning models, we implemented MobileNetV1 \citep{howard2017mobilenets}, VGG19p \cite{simonyan2014very}, MobileNetV2 \citep{howard2018inverted}, ResNet50 \citep{he2016deep}, and EfficientNet \citep{tan2019efficientnet} architectures from scratch without using pre-trained weights. The models were trained using the Adam optimizer with a learning rate of $1 \times 10^{-3}$, batch size of 512, and 15 epochs. We used mean squared error (MSE) as the loss function for coordinate prediction tasks and categorical cross-entropy for building and floor classification. No additional regularization techniques were applied beyond the architectural regularization inherent in the selected models.

\subsubsection{Conformal Prediction Implementation}
We implemented the standard Split Conformal Prediction approach. The procedure consisted of these steps:
\begin{enumerate}
    \item Train the model on the training dataset (70\% of the data).
    \item For each sample in the calibration set (10\% of the data), compute the non-conformity score defined as the Euclidean distance between the predicted coordinates and the ground truth coordinates: $s_i = \|Y_i - \hat{Y}_i\|_2$.
    \item Determine the quantile threshold $\hat{q}$ for a given significance level $\alpha$ using the empirical distribution of the calibration scores.
    \item For each test sample, construct the prediction set by including all candidate positions whose non-conformity scores are below the threshold $\hat{q}$.
    \item Evaluate coverage (the percentage of test samples where the true position is contained in the prediction set) across different significance levels $\alpha$.
\end{enumerate}

This procedure allows us to provide statistical guarantees on the correctness coverage while quantifying the uncertainty of the positioning system through the size of the prediction sets.
\begin{figure}
\centering
\includegraphics[width=0.95\linewidth]{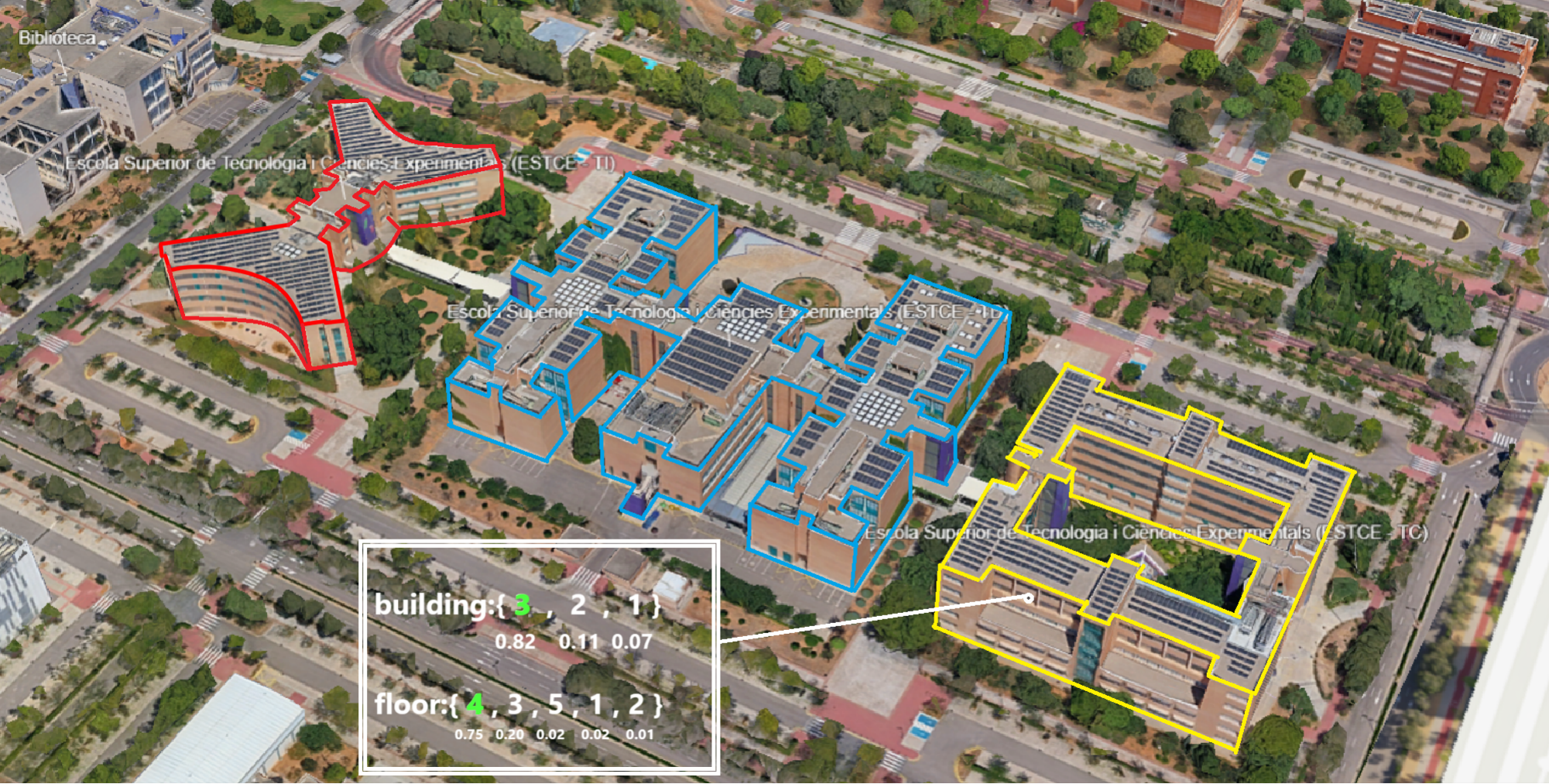}
\caption{RSSI feature classification.
\label{fig1}}
\end{figure}   
As previously mentioned, conformal prediction can construct prediction sets for positioning problems, allowing their coverage to be artificially controlled. Additionally, the size of the prediction set reflects the model's performance and indicates the degree of uncertainty. Path planning involves using the lambda threshold of risk control to identify the most reliable points for each path. During the calibration phase, the most appropriate threshold is determined to ensure that the intersection between the selected points and the actual path achieves a ratio no less than $1-\alpha$, thereby ensuring that the loss is no more than $\alpha$. The schematic diagrams of feature classification and path positioning are presented in Figure~\ref{fig1} and Figure~\ref{fig2}.
\begin{figure}
\centering
\includegraphics[width=0.65\linewidth]{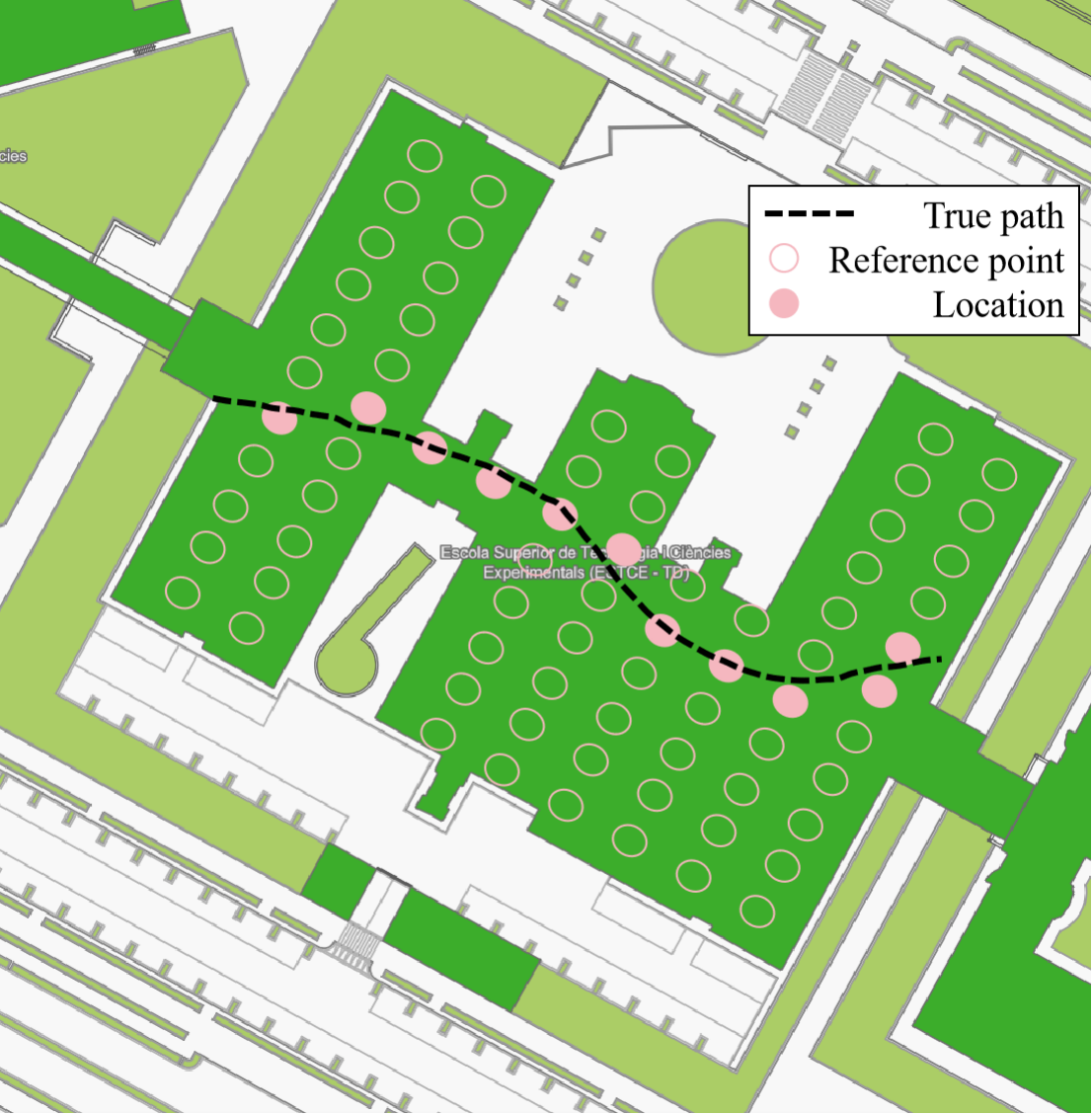}
\caption{Path positioning.
\label{fig2}}
\end{figure}   
As shown in Figure~\ref{fig3_1} and Figure~\ref{fig3_2}, the experimental results using a fully connected CNN on this dataset indicate that as the number of training rounds increases, the loss decreases, and the accuracy improves. Although the model performs well on the training set, its generalization ability on the test set is limited, with the test accuracy eventually stabilizing at 0.85.

Using the conformal prediction framework discussed above and considering the practical application of indoor positioning, we selected five lightweight models: MobileNetV1, VGG19, MobileNetV2, ResNet50, and EfficientNet. The experimental results are presented in Figure~\ref{fig4} and Figure~\ref{fig5}.

Figure~\ref{fig4} illustrates the variation in empirical coverage and target coverage of each model as a function of alpha. A higher alpha decreases the likelihood that the prediction set will contain the correct category, leading to a lower accuracy in the positioning results. Overall, the empirical coverage curve of each model closely aligns with the target coverage curve, demonstrating that the conformal prediction technique effectively approximates the target coverage in these deep learning models. Despite the theoretical guarantee of SCP being rigorous, there can be minor fluctuations in practice due to finite-sample variability. 
In real-world tasks of indoor positioning, test points may deviate from the distribution of the calibration set, and at this point, there will be miscalibration issues of the coverage occurring. 

Figure~\ref{fig5} shows the variation in the average prediction set size with alpha, where the prediction set size for each model exhibits a decreasing trend. This indicates that reducing the coverage requirements results in smaller prediction set sizes. Among the models, VGG19 has a relatively larger prediction set size at the same alpha, suggesting that the prediction set generated by VGG19 is broader and exhibits greater uncertainty when quantified. In contrast, EfficientNet generates a relatively smaller prediction set size and demonstrates superior performance in indoor localization tasks. Specific data are shown in Table~\ref{prediction_set_size_data}.

\begin{figure}[t]
\centering
\includegraphics[width=0.7\linewidth]{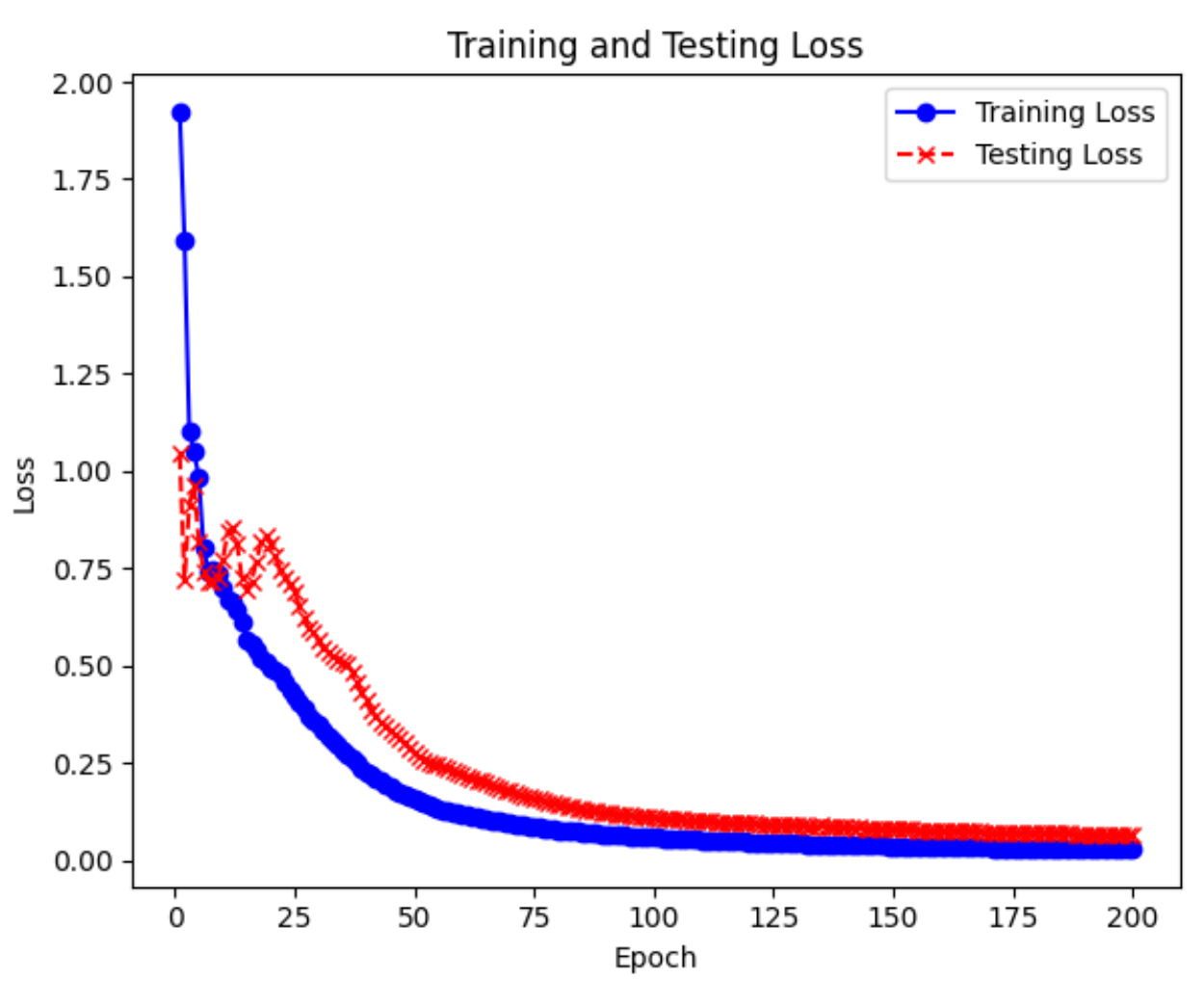}
\caption{Training and testing loss.
\label{fig3_1}}
\end{figure} 

\begin{figure}[t]
\centering
\includegraphics[width=0.7\linewidth]{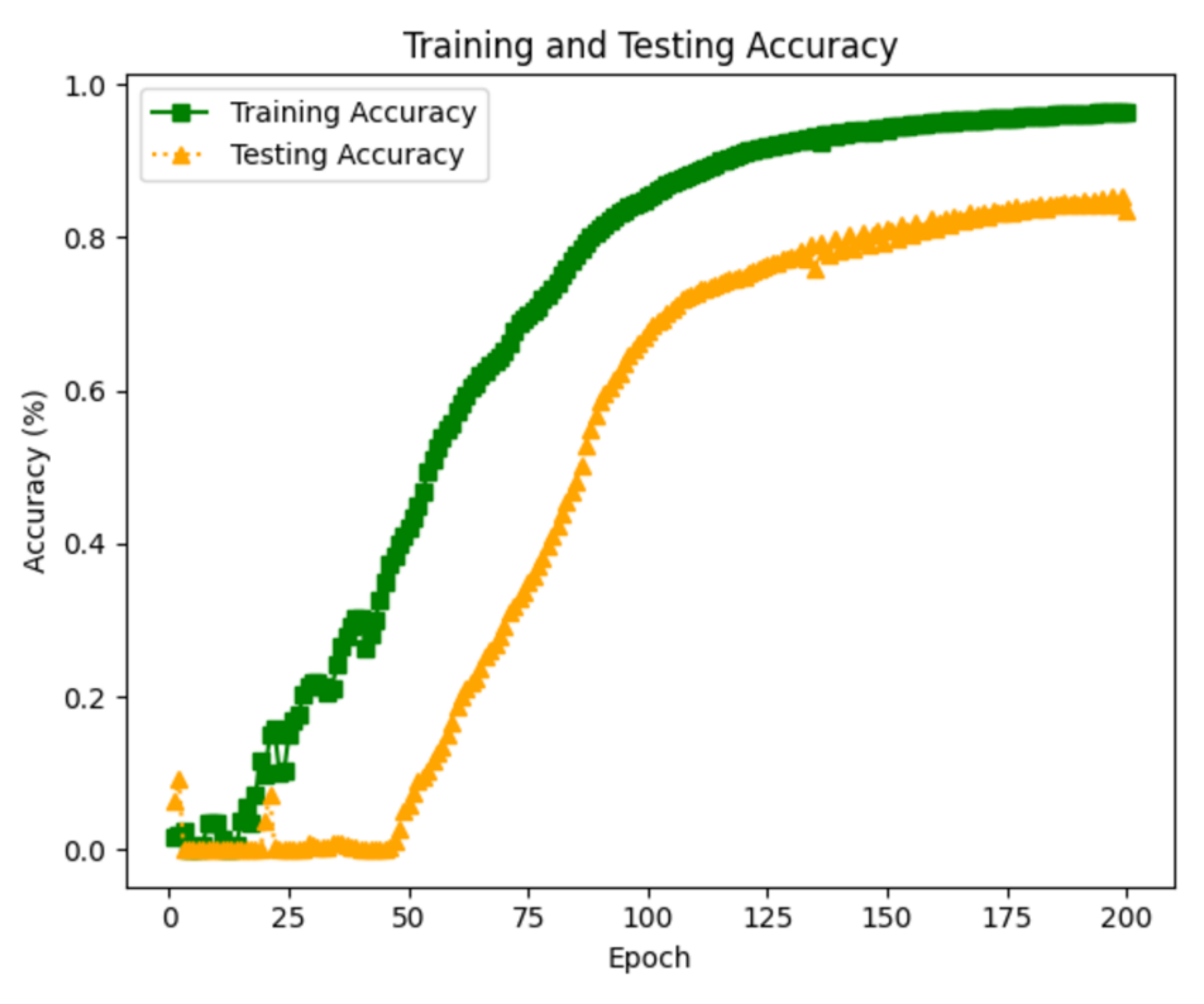}
\caption{Training and testing accuracy.
\label{fig3_2}}
\end{figure}   

\begin{figure}[t]
\centering
\includegraphics[width=0.7\linewidth]{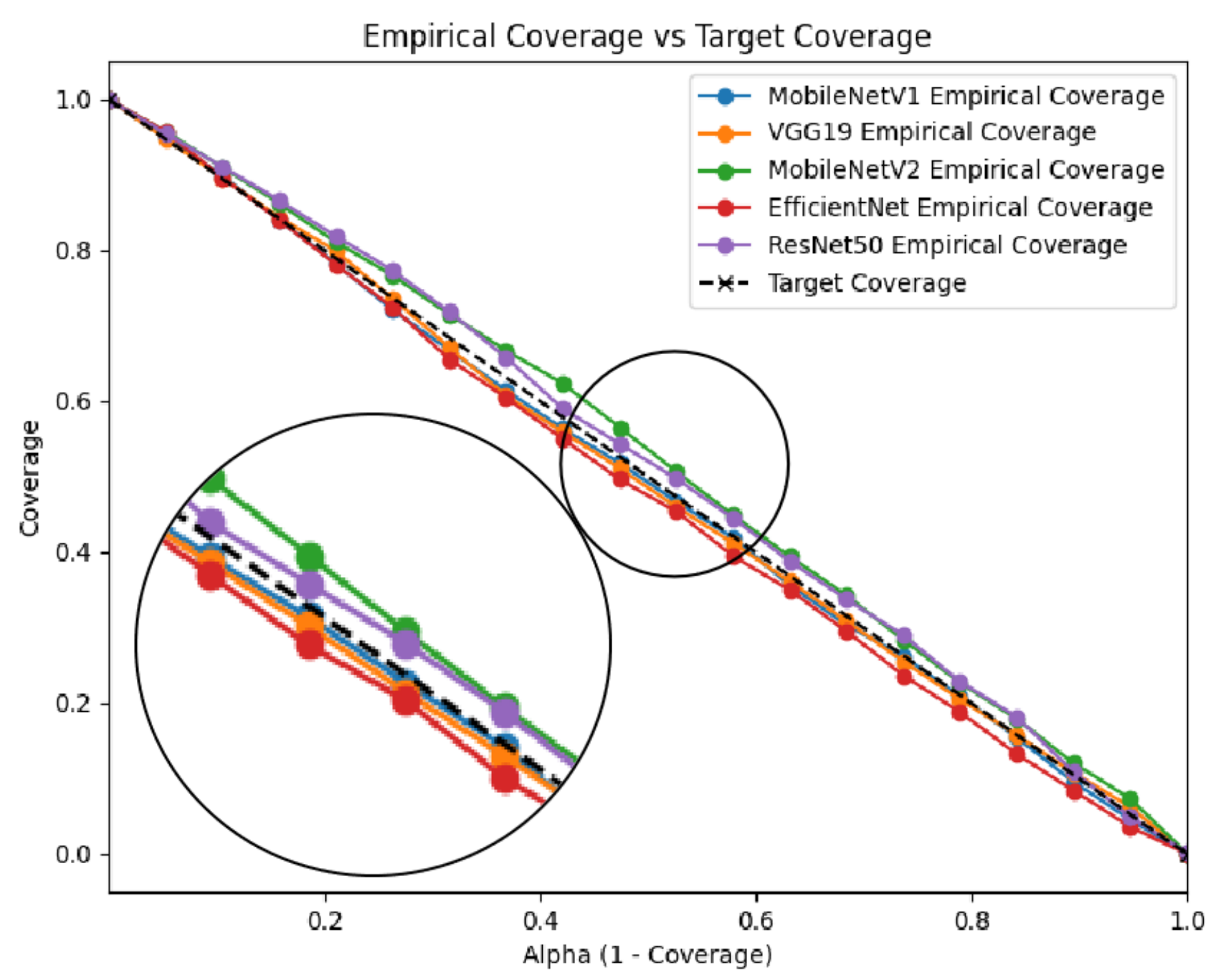}
\caption{Empirical and target coverage.
\label{fig4}}
\end{figure} 

\begin{figure}[t]
\centering
\includegraphics[width=0.7\linewidth]{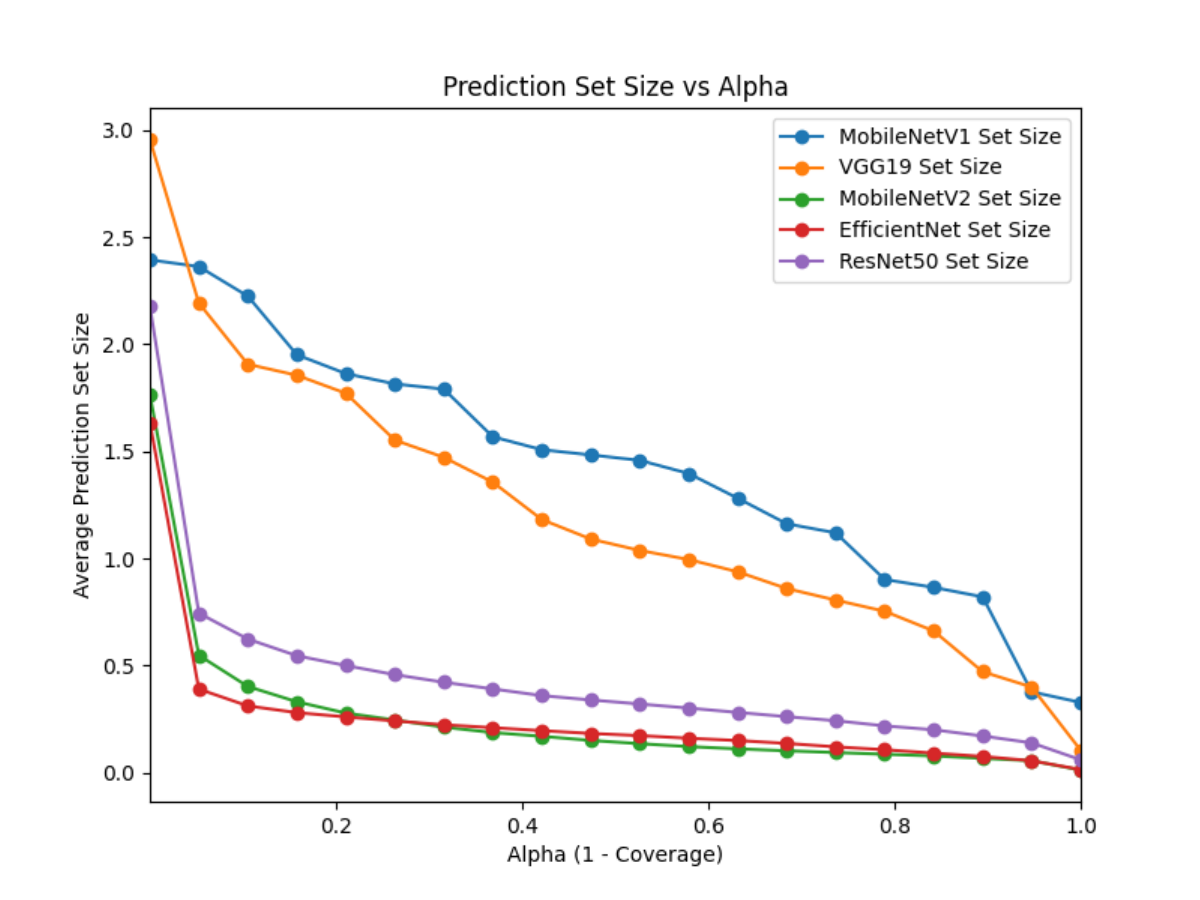}
\caption{Average prediction set size.
\label{fig5}}
\end{figure}   

\begin{table}[t]
  \caption{Average prediction set size \label{prediction_set_size_data}}
  \centering
  \begin{tabular}{lllll}
    \toprule
    \multicolumn{5}{c}{Models}                   \\
    \cmidrule(r){2-5}
    \textbf{$Alpha$} & \textbf{MobileNetV1}	& \textbf{VGG19}	& \textbf{MobileNetV2}	& \textbf{ResNet50}\\
    \midrule
    0 & 2.342 & 2.86 & 1.76 & 2.14\\
    0.052 & 2.311 & 2.15 & 0.63 & 0.818\\
    0.105 & 2.191 & 1.89 & 0.505 & 0.709\\
    0.158 & 1.934 & 1.84 & 0.432 & 0.636\\
    0.211 & 1.853 & 1.77 & 0.396 & 0.599\\
    0.263 & 1.802 & 1.57 & 0.354 & 0.557\\
    0.316 & 1.789 & 1.49 & 0.328 & 0.526\\
    0.368 & 1.576 & 1.39 & 0.302 & 0.495\\
    0.421 & 1.519 & 1.22 & 0.286 & 0.469\\
    0.474 & 1.5 & 1.14 & 0.27 & 0.453\\
    0.526 & 1.475 & 1.09 & 0.26 & 0.427\\
    0.579 & 1.412 & 1.05 & 0.244 & 0.406\\
    0.632 & 1.311 & 0.996 & 0.229 & 0.39\\
    0.684 & 1.198 & 0.927 & 0.229 & 0.375\\
    0.737 & 1.16 & 0.877 & 0.218 & 0.359\\
    0.789 & 0.96 & 0.827 & 0.213 & 0.328\\
    0.842 & 0.922 & 0.745 & 0.197 & 0.312\\
    0.895 & 0.877 & 0.57 & 0.187 & 0.296\\
    0.947 & 0.464 & 0.501 & 0.176 & 0.255\\
    1 & 0.426 & 0.225 & 0.135 & 0.182\\
    \bottomrule
  \end{tabular}
\end{table}

\subsection{Model Architecture Analysis and Performance Differences}
The observed differences in prediction set sizes among the various architectures reveal important insights about their uncertainty quantification capabilities, with EfficientNet consistently producing smaller prediction sets while maintaining the target coverage, indicating higher certainty in its predictions. 

This superior performance stems from several architectural advantages, including its compound scaling strategy, which uniformly scales network width, depth, and resolution using fixed coefficients, leading to more efficient feature extraction and reduced uncertainty compared to models that scale only one dimension. Additionally, EfficientNet’s mobile inverted bottleneck convolution (MBConv) blocks, enhanced with squeeze-and-excitation optimization, allow the model to focus on the most discriminative features in WiFi fingerprints, which is particularly valuable in indoor positioning where certain access points provide more informative signals. Furthermore, despite its strong performance, EfficientNet maintains parameter efficiency with a relatively small parameter count compared to models like VGG19, reducing overfitting risks and improving generalization, which ultimately results in tighter, more reliable prediction sets that better reflect true uncertainty.

In contrast, VGG19's larger prediction sets reflect greater uncertainty in its positioning estimates, which can be attributed to several architectural limitations. First, its uniform architecture lacks modern bottleneck designs, following instead a simple pattern of increasing channel width with depth. This approach may cause VGG19 to capture both useful signals and noise from WiFi fingerprints, unlike models such as EfficientNet and ResNet50, which focus more efficiently on essential features. Additionally, VGG19 lacks explicit attention mechanisms, preventing it from prioritizing the most relevant access points for positioning. As a result, its confidence may be diluted across multiple possible locations. Finally, its large parameter count increases overfitting risks, particularly in dynamic indoor environments where test samples often deviate slightly from the training distribution, leading to higher uncertainty in real-world deployments.

MobileNetV1 and MobileNetV2 both achieve a balance between parameter efficiency and prediction certainty, with MobileNetV2 showing slightly better performance due to its inverted residual structure. ResNet50, despite its depth, performs reasonably well but does not match EfficientNet's efficiency in uncertainty quantification.

\subsection{Connecting Experimental Results to Theoretical Framework}
Our experimental results provide empirical validation of the theoretical guarantees established in the methods section. The observed alignment between target and empirical coverage across all models (Figure~\ref{fig4}) confirms the fundamental property of conformal prediction stated in Equation \ref{1}:

\begin{equation}
1 - \alpha \leq \mathbb{P}(Y_{\text{test}} \in C(X_{\text{test}})) \leq 1 - \alpha + \frac{1}{n + 1}
\end{equation}

This empirical validation is significant, as it demonstrates that the theoretical coverage guarantees hold regardless of the underlying model architecture, affirming the model-agnostic nature of conformal prediction. The minor fluctuations observed in the coverage curves can be attributed to the finite size of the calibration set ($n$), which introduces a small margin of error ($\frac{1}{n+1}$) as predicted by the theory.

The prediction set size analysis (Figure~\ref{fig5}) illustrates the practical implications of our theoretical framework for uncertainty quantification. As established in our non-conformity score design discussion, the size of the prediction set directly reflects the model's uncertainty. The decreasing trend in set size as alpha increases demonstrates the trade-off between coverage guarantee and precision that system designers can control.

The differences in prediction set sizes among models at the same alpha level highlight an important aspect of our theoretical framework: while conformal prediction guarantees coverage regardless of the model, the efficiency of this coverage (as measured by prediction set size) depends on the accuracy and calibration of the underlying model. This connects to our discussion of non-conformity scores in Section 3.1, where we emphasized that the choice of model and scoring function affects the shape and size of prediction sets while maintaining the coverage guarantee.

Furthermore, these results validate our approach of using Euclidean distance as the non-conformity score for positioning tasks. The significant differences in the sizes of the prediction set in all models confirm that this distance-based score effectively captures the physical reality of positioning errors, making it particularly suitable for indoor positioning applications where spatial relationships are important.
%%%%%%%%%%%%%%%%%%%%%%%%%%%%%%%%%%%%%%%%%%
\section{Conclusions}
In this study, we have effectively tackled the challenges related to deep learning in indoor positioning. By applying conformal prediction, we have provided a novel approach to ensuring the reliability and security of positioning decisions, along with uncertainty quantification. The introduced conformal risk control and conformal p-value framework have enhanced the performance of indoor positioning in path navigation tasks.

The experimental results on the UJIIndoLoc dataset validate the effectiveness of the proposed methods. The conformal prediction technique can accurately approximate the target coverage in deep learning models, which is a significant improvement over traditional methods. The analysis of different lightweight models shows that they exhibit varying levels of performance in terms of prediction set size and uncertainty quantification. For example, EfficientNet performs better in indoor localization tasks with a relatively smaller prediction set size, while VGG19 has a larger prediction set size, indicating greater uncertainty.

In the future, we intend to further investigate the use of conformal prediction in more complex indoor environments and with a broader range of sensors. Moreover, we aim to refine the existing algorithms to decrease computational complexity and enhance real-time performance, making the indoor positioning system better suited for resource-limited and time-sensitive applications.

%%%%%%%%%%%%%%%%%%%%%%%%%%%%%%%%%%%%%%%%%%

\begin{ack}
This research was funded by the National Training Program of Innovation and Entrepreneurship for Undergraduates (202410635117).
\end{ack}

% \section*{References}
\bibliographystyle{unsrtnat}
\bibliography{references}
\end{document}